%% file: main.tex
\documentclass[10pt,twocolumn,letterpaper]{article}

\usepackage{cvpr}              


\definecolor{cvprblue}{rgb}{0.21,0.49,0.74}
\usepackage[pagebackref,breaklinks,colorlinks,allcolors=cvprblue]{hyperref}
\usepackage{url}
\usepackage{algorithm}
\usepackage{algorithmic}
\usepackage{array}
\usepackage{amssymb}
\usepackage{colortbl}
\definecolor{shadecolor}{rgb}{0.92,0.92,0.92}  
\usepackage{framed} 
\definecolor{myColor}{HTML}{ffe5f0}
\newcommand{\myCellColor}[1]{\cellcolor{myColor}#1}
\usepackage{tikz}
\usepackage{bm}

\usepackage{tcolorbox}
\usepackage{booktabs} 
\usepackage{multirow}
\usepackage{makecell}

\def\paperID{3996} 
\def\confName{CVPR}
\def\confYear{2025}

\title{C-Drag: Chain-of-Thought Driven Motion Controller for Video Generation}

\author{Yuhao Li$^{1,2}$, Mirana Claire Angel$^3$,  Salman Khan$^{1,5}$ Yu Zhu$^1$, Jinqiu Sun$^1$, \\Yanning Zhang$^1$,Fahad Shahbaz Khan$^{2,4}$\\
$^1$Northwestern Polytechnical University\\
$^2$Mohamed bin Zayed University of Artificial Intelligence\\
$^3$9009.ai~~~$^4$Linköping University
$^5$Australian National University
}

\begin{document}

\twocolumn[{
\renewcommand\twocolumn[1][]{#1}
\maketitle
\begin{center}
\captionsetup{type=figure}
\includegraphics[width=\textwidth]{ 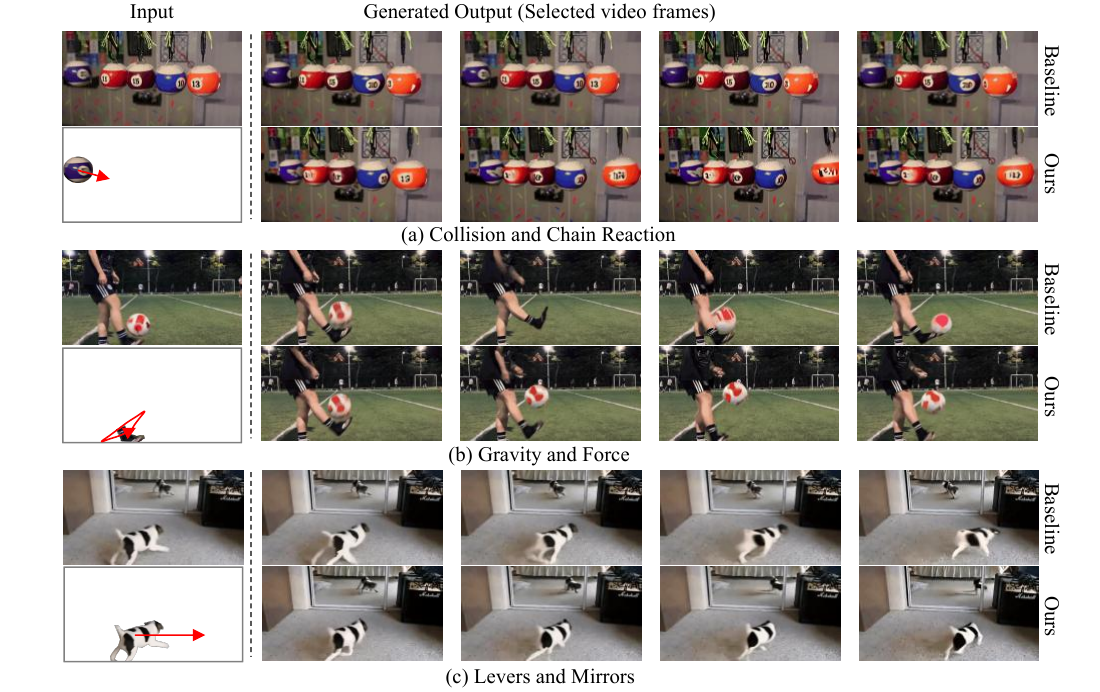}
\captionof{figure}{Our C-Drag employs a single trajectory control signal (\textcolor{red}{red} arrow), integrated with a vision-language model (VLM) and Chain-of-Thought (CoT) reasoning, to generate controllable videos that emphasize motion realism. Results are illustrated in three example scenarios, each comprising two rows: baseline output (\emph{top}) and C-Drag output (\emph{bottom}). \textbf{(a)} Collision and Chain Reaction: The trajectory of a single sphere leads to complex collisions and chain reactions among multiple spheres. \textbf{(b)}  Gravity and Force: A foot’s trajectory impacts a football, showing motion under gravitational and force dynamics. \textbf{(c)}  Levers and Mirrors: A puppy's movement is reflected in a mirror, showcasing coupled motion control through mirror reflection. Best viewed zoomed in. Additional results are presented in suppl. material.
}
\label{fig_Intro}
\end{center}
}]

\begin{abstract}

Trajectory-based motion control has emerged as an intuitive and efficient approach for controllable video generation. However, the existing trajectory-based approaches are usually limited to only generating the motion trajectory of the controlled object and ignoring the dynamic interactions between the controlled object and its surroundings. To address this limitation, we propose a Chain-of-Thought-based motion controller for controllable video generation, named C-Drag. Instead of directly generating the motion of some objects, 
our C-Drag first performs object perception and then reasons the dynamic interactions between different objects according to the given motion control of the objects. Specifically, our method includes an object perception module and a Chain-of-Thought-based motion reasoning module. The object perception module employs visual language models to capture the position and category information of various objects within the image. The Chain-of-Thought-based motion reasoning module takes this information as input and conducts a stage-wise reasoning process to generate motion trajectories for each of the affected objects, which are subsequently fed to the diffusion model for video synthesis.  
Furthermore, we introduce a new video object interaction (VOI) dataset to evaluate the generation quality of motion controlled video generation methods. Our VOI dataset contains three typical types of interactions and provides the motion trajectories of objects that can be used for accurate performance evaluation.  Experimental results show that C-Drag achieves promising performance across multiple metrics, excelling in object motion control. 
Our benchmark, codes, and models will be  available at ~\url{https://github.com/WesLee88524/C-Drag-Official-Repo}.

\end{abstract}
\vspace{-20pt}
\section{Introduction}

Controllable video generation~\cite{lotter2016deep, srivastava2015unsupervised, chiappa2016recurrent, wu2021godiva, wu2022nuwa,  ho2022imagen, singermake, yin2023nuwaxl, esser2023structure} has made significant progress in recent years, largely due to advances in diffusion models. 
Current approaches fall into three categories: text-based, image-conditioned, and trajectory-based video generation.
Text-based approaches~\cite{chen2023control, zhang2024fastvideoedit, an2023latent, wu2021godiva, wu2022nuwa,  singermake, ho2022imagen, zhang2023controlvideo} aim to generate videos according to given text descriptions, while image conditional approaches~\cite{hu2022make, yin2023nuwaxl, esser2023structure, ni2023condi, niu2024mofa, hu2024animateanyone} generate videos that are similar to the style of the given image. 
Although these approaches can perform controllable video generation via text or image, they cannot precisely control the object motion across frames. 
In contrast, trajectory-based approaches focus on building object motion across frames~\cite{qiu2024freetraj, ma2023trailblazer, zhang2024tora, wu2024draganything, yin2023dragnuwa, li2024dragapart, zhang2024dragentity}. However, most trajectory-based approaches face challenges in handling multi-object interactions in complex scenes.

\begin{figure}[t]
    \centering
    \includegraphics[width=1\linewidth]{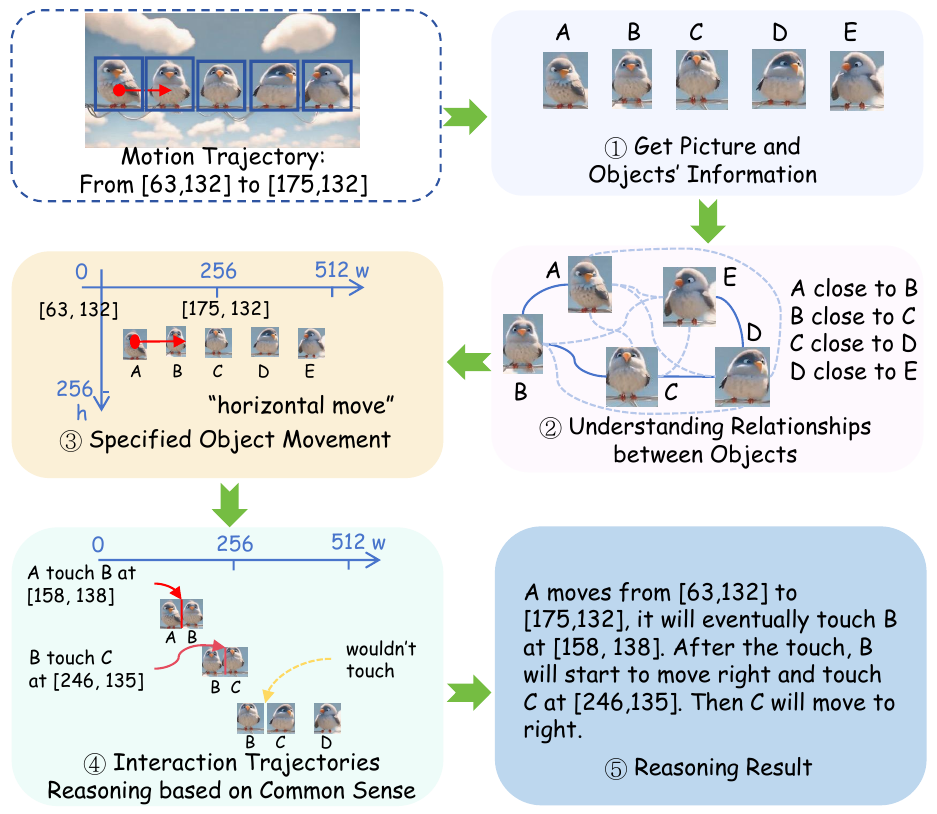}
    \caption{Our C-Drag approach is motivated from human cognitive patterns to model dynamic interactions between objects for controllable video generation. Human reasoning about object interactions involves few key stages. First, obtaining information about the image and objects. Next, inferring relationships between objects. Then, based on the trajectory of a specific object and motion principles, predicting the reactions of other objects.  Finally, determining the overall result of these interactions. }
    \label{fig:Chain-of-Thought}
\vspace{-6pt}
\end{figure}

Human cognitive patterns~\cite{kahneman1992reviewing,goodale1992separate,ungerleider1994and, van2010retrieval} suggest that motion reasoning, particularly for multi-object interactions, requires two key elements: object perception and cognitive ability.
First, accurate content perception involves identifying and locating all relevant objects.
Second, reasoning requires cognitive skills, including common sense knowledge linking visual information to real-world dynamics, such as collisions triggering chain reactions or synchronous motion of an object and its reflection. 
We argue that the typical human reasoning process for motion trajectory prediction of all objects in scenes typically involves several stages, as illustrated in Figure~\ref{fig:Chain-of-Thought}: (1) accurately identifying and locating all objects, such as the five birds and the wire; (2) comprehending their relationships, such as the arrangement of five birds sitting side by side on the wire from left to right; (3) locating a specific bird (the leftmost one) based on given trajectory and clarifying its future movement path; (4) 
 inferring potential interactions, such as a collision between this bird and others potentially triggering a chain reaction; and
(5) ensuring that the reasoning aligns with coherent cognitive behavior principles.

Recently, vision-language models (VLMs) have shown impressive visual understanding and reasoning capabilities~\cite{zhang2023video, maaz2023video, lin2023video}. At the same time, Chain-of-Thought (CoT) prompting~\cite{wei2022chain}  has gained attention for enhancing large language models (LLMs) reasoning ability.
CoT intuitively decomposes complex problems into simpler manageable sub-problems, promoting a human-like reasoning. While this approach has proven effective in language understanding and reasoning tasks~\cite{li2023videochat, fei2024video}, CoT-based reasoning framework specifically built for controllable video generation with trajectory inputs remains unexplored.

In this work we introduce C-Drag, a Chain-of-Thought driven motion controller for video generation, which effectively leverages the perception and reasoning abilities of vision-language models (VLMs). Our C-Drag contains an object perception module and a CoT-based motion reasoning module. The object perception module first utilizes class-agnostic segmentation to capture the controlled object accurately and then uses a VLM to detect all objects within the image. 
The CoT-based reasoning module introduces a stage-wise reasoning strategy to precisely reason motion trajectories of all objects according to the detected position and category information. With the generated object trajectories, we employ a trajectory-based generation model to generate the videos with multiple-object interactions.
Furthermore, we build a new video object interaction  (VOI) benchmark to evaluate the performance of motion controlled video generation methods. Our VOI dataset has 72 videos and contains three types of object interactions, including \textit{collision and chain reaction}, \textit{gravity and force}, and \textit{levers and mirrors}. We also provide the ground-truth trajectories of objects across frames and introduce a metric, MOC, for performance evaluation that represents the plausibility and realism of object motion by calculating the similarity between predicted and ground-truth trajectories.

\noindent\textbf{Contributions.}
We introduce C-Drag, a Chain-of-Thought driven motion controller for video generation. C-Drag strives to reason precise motion trajectories according to object interactions by an object perception module and a Chain-of-Thought-based motion reasoning module. 
We build a new video object interaction (VOI) dataset for evaluating trajectory-based video generation. Our VOI dataset contains three typical motion types and provides object motion  trajectories. 
C-Drag can generate high-quality videos according to the given object motion control (see Figure~\ref{fig_Intro}), outperforming the baseline by around 35.5\% in terms of  MOC score on the VOI dataset.

\section{Related Works}
\noindent \textbf{Controllable Video Diffusion.}
Recently, diffusion models have achieved great progress on video synthesis, which are usually built on image generation approaches.
ControlNet \cite{zhang2023controlnet} introduces a novel architecture to integrate spatial conditioning controls into pre-trained text-to-image diffusion models. 
GLIGEN \cite{li2023gligen} designs a gated mechanism to incorporate  grounding information for image generation. Afterwards, many video models have been proposed which aim to generate controllable videos according to user-specified conditions.
Control-A-Video \cite{chen2023control}  leverages both the content prior and motion prior for video generation, and supports  video generation according to different control signals.
Imagen Video \cite{ho2022imagen} first generates videos using a base video generation model, and then cascades spatial and temporal video super-resolution models to generate high-definition videos.
Video LDM \cite{rombach2021highresolution} extends image generation model to video by fine-tuning temporal layers. Text2Video-Zero \cite{khachatryan2023text2video} introduces post-processing techniques to maintain temporal consistency for zero-shot generation.

In addition, few works explore using the reference videos or skeleton poses as input for video generation. 
MotionEditor \cite{tu2024motioneditor} integrates a content-aware motion adapter into ControlNet for video motion editing.
UniAnimate \cite{wang2024unianimate} presents a framework to deal with a reference image and a noised video to improve feature alignment. 
Despite these efforts, designing a method to generate desired motion while preserving spatio-temporal dynamics is still an open problem.

\noindent \textbf{Trajectory-oriented Diffusion.}
Trajectory-based methods have gained great attention for their user-friendliness and the ability to control object motion effectively. DragAnything~\cite{wu2024draganything} employs an entity representation for entity-level motion control of video generation.
Tora \cite{zhang2024tora} introduces a trajectory-oriented diffusion transformer along with a trajectory extractor and a motion-guidance fuser. 
These approaches present the initially potential ability
in simulating real-world movements. However, they usually focus on generating the motion only for pre-defined objects, which do not have comprehensive understanding of the scene or consider the complex interactions of objects.

\noindent \textbf{Grounding and Reasoning with LLMs.}
Large Language Models (LLMs) have shown promising capabilities in grounding and reasoning tasks. 
VideoDirectorGPT \cite{lin2023videodirectorgpt} explores using LLMs for  multi-component video plan, which is used to guide video generation. 
LLM-grounded Video Diffusion (LVD) \cite{lian2023llmgroundedvideo} first leverages LLMs to generate dynamic scene layouts,  and then uses these layouts to guide video diffusion models. Video-of-Thought~\cite{fei2024video}  introduces a reasoning framework for fine-grained spatial-temporal video grounding, which develops step-by-step Chain-of-Thought (CoT) \cite{wei2022chain}. While Video-of-Thought~\cite{fei2024video} employs a step-by-step CoT-based reasoning (e.g., object tracking, action analysis) for video question answering, our work explores reasoning capabilities for video generation task.
Further, our work not only requires the ability to classify the motion type of the object by analyzing a single frame image, but also desires the inference of the future trajectory of the object based on the single frame image thereby treating it as a regression problem. Our approach strives to encode dynamic behavior of objects in a single image to better cope with different motion patterns.

PhysGen \cite{liu2024physgen} uses LLMs to infer the physical parameters of rigid objects and employs a physics simulator to generate realistic object behavior videos. 
However, it struggles with deformable objects and accurately estimating physical parameters from visual input, leading to discrepancies between the generated videos and real-world dynamics. Moreover, its generalization ability is limited as it can only simulate planar motion resulting in sub-optimal results in complex and diverse scenarios. 

\noindent \textbf{Our Approach.}
Different from existing works, we propose a training-free Chain-of-Thought driven motion controller for video generation. By leveraging the capabilities of visual language models (VLM) in combination with a pre-trained trajectory-based video generation model, our approach effectively addresses the aforementioned limitations. In object interaction scenarios, our CoT-based motion reasoning module gradually decomposes complex multi-object interactions to capture causal relationships and interaction forces, thereby allowing consistent trajectory generation without needing a high-precision physics engine.

\section{Method}

\begin{figure}[t]
    \centering
    \includegraphics[width=1\linewidth]{ 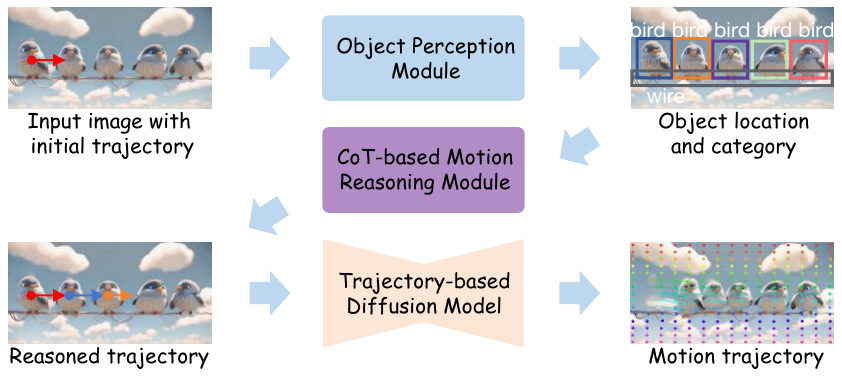}
    \caption{Overview of our C-Drag. C-Drag first takes a single RGB image and one or more drag motion trajectories as input. We employ an object perception module to obtain information about all related objects in the image. Chain-of-Thought (CoT)-based reasoning module introduces a reasoning strategy to precisely reason motion trajectories of all objects according to the detected position and category information. With the generated object trajectories, we use a pre-trained trajectory-based generation model to generate the videos with multiple-object interactions. }
    \label{fig:Pipeline}
\vspace{-12pt}
\end{figure}

\textbf{Motivation.} Recently, trajectory-based video generation has gained attention due to its user-friendliness and precise control. However, maintaining visual consistency and logical coherence across frames remains a challenge, especially in complex scenes involving multiple interacting objects. This limitation hinders its application in high-precision tasks.
In contrast, humans can successfully deal with such scenarios by decomposing the process into several stages, as illustrated in Figure~\ref{fig:Chain-of-Thought}. Specifically, we begin by observing the scene, locating objects, and understanding the relationships between different objects. Afterwards, we predict the interactions of objects. Finally, we estimate the trajectories of objects.
Based on this, we introduce C-Drag, a  Chain-of-Thought driven motion controller for video generation. By decomposing complex multi-object interactions into sequential reasoning stages, C-Drag generates high-quality videos that maintain visual coherence and realistic consistency.

\noindent\textbf{Overall Architecture.} Figure~\ref{fig:Pipeline} shows the overall architecture of our proposed C-Drag. It mainly consists of three components, including an object perception module, a CoT-based motion reasoning module and a trajectory-based video generation module. Given the input image and corresponding motion trajectory, the object perception module employs class-agnostic segmentation and open-set detection
to obtain the object location, mask, and category information (Sec. \ref{sec:opm}). Based on this object information, our CoT-based motion reasoning module predicts the motion trajectories of objects according to stage-wise reasoning (Sec. \ref{sec:cot}). These generated object trajectories are fed to a pre-trained trajectory-based video generation model to generate high-quality videos. Next, we present the object perception and CoT-based motion reasoning modules.

\begin{figure*}[t]
    \centering
    \includegraphics[width=1\linewidth]{ 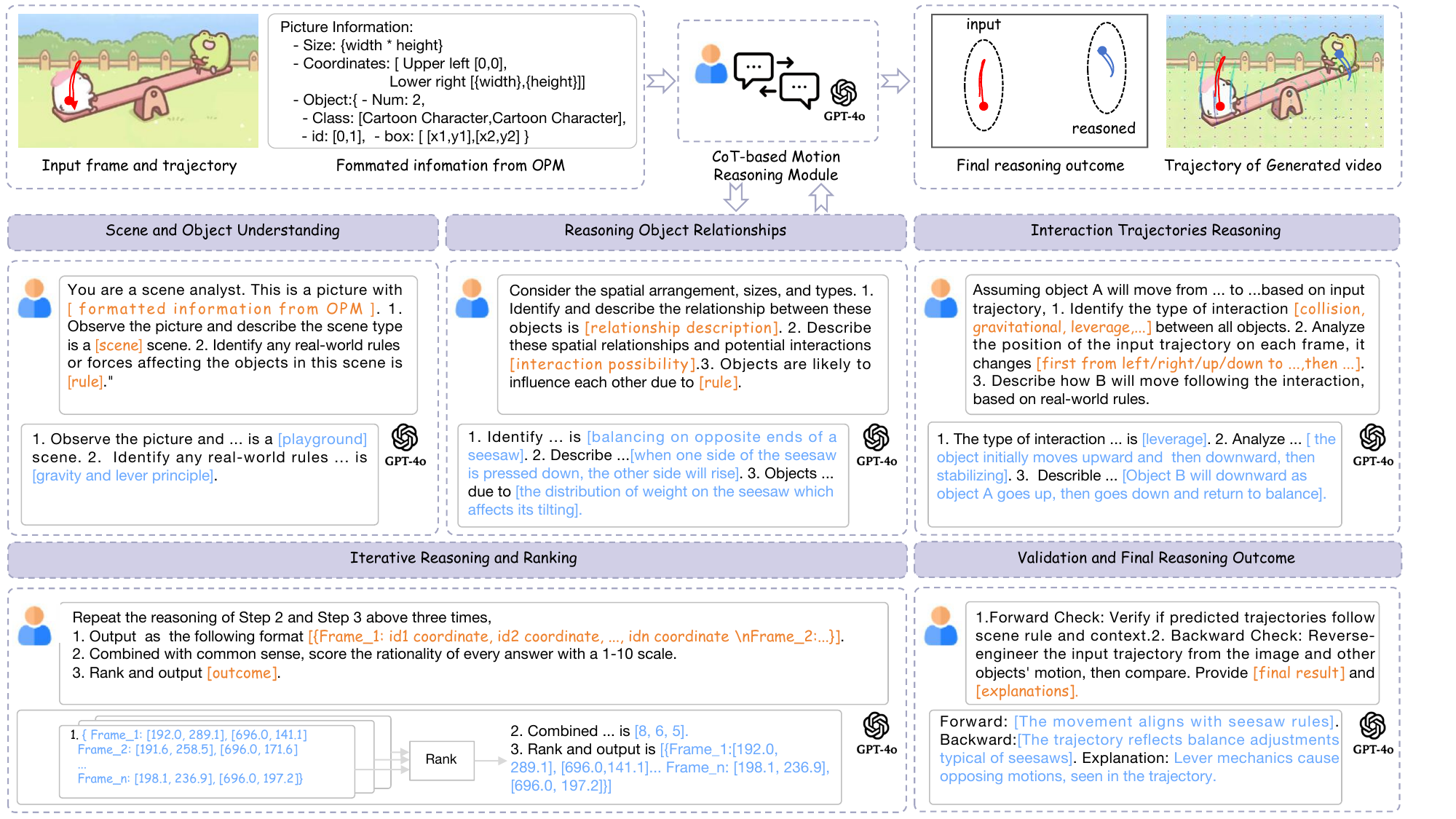} 
    \caption{An illustrative view of CoT-based Motion Reasoning Module which undergoes a five-stage reasoning process. \textbf{Scene  and Object Understanding}, where a pre-trained visual language model (VLM) interprets the scene and establishes motion rules using formated information from Object Perspection Module. In \textbf{ Reasoning Object Relationship}, the VLM identifies spatial relationships and potential interactions among objects to inform trajectory predictions. \textbf{ Interaction Trajectories Reasoning} follows, categorizing interactions (e.g., collisions, forces) and predicting affected object paths. During \textbf{Iterative Reasoning and Ranking}, initial predictions are iteratively optimized, with the VLM selecting the most consistent motion sequences. Finally, in \textbf{Validation and Final Reasoning Outcome}, forward and backward validation ensures predicted trajectories align with scene rules, iterating until accuracy is achieved.}
    \label{fig:CoT-detailed}
    \vspace{-12pt}
\end{figure*}

\subsection{Object Perception Module}\label{sec:opm}
Object perception module aims to extract the information, including location and category, of all objects in input image. For the input image  \( \bm{I} \in \mathbb{R}^{H \times W \times 3} \), we first employ a class-agnostic segmentation model~\cite{kirillov2023segment}
to generate the mask $\bm{M}_{\text{c}}$ of controlled object  corresponding to the beginning point $\bm{P}_{b}$ of   motion trajectory as
\begin{equation} 
\bm{M}_{\text{c}} = \text{Segment}(\bm{I}, \bm{P}_{b}).
\end{equation}
Then, we feed the mask of controlled object $\bm{M}_{\text{c}}$ and a pre-defined text prompt $T$ to the VLM
to detect all the objects,  represented as \( \mathcal{O} = \{(\bm{B}_i, \bm{C}_i)\}_{i=1}^{N} \), where \( \bm{B}_i = (x_{i}^1, y_{i}^1, x_{i}^2, y_{i}^2) \) are the coordinates of bounding box, \( \bm{C}_i \) is the object category, and $N$ is the number of objects. 
Afterwards, we  employ an open-set detection model~\cite{liu2023grounding}
to refine the bounding-boxes and generate the masks as
\begin{equation} 
(\bm{B}^{f}, \bm{M}^{f}) = \text{Open-set Detection}(\bm{I}, \mathcal{O}). 
\end{equation}
Finally, we generate the output information of all objects as  \( \mathcal{O}_{\text{final}} = \{(\bm{B}_{i}^f, \bm{C}_i, \bm{M}_i^{f})\}_{i=1}^{N} \), where \( \bm{B}_{i}^f \) represents the refined bounding-box coordinate of object $i$, and \(  \bm{M}_i^{f} \) is the fine-grained segmentation masks of object $i$.

The object perception module extracts object-aware information, which is forwarded to the subsequent processing stage. Based on this information, our CoT-based motion reasoning module then effectively predicts the trajectories of the controlled object and other objects affected by it.

\subsection{CoT-based Motion Reasoning Module}\label{sec:cot}
We introduce a CoT-based motion reasoning module that decomposes trajectory prediction into multiple stages
which allows for a more detailed and structured reasoning process, thereby enhancing the model's scene understanding ability and planning performance.
Figure~\ref{fig:CoT-detailed} presents the details of our CoT-based motion reasoning module. At each stage, we design a text prompt and feed the text prompt to the VLM model 
for reasoning.

\noindent\textbf{Scene and Object Understanding.} This stage aims to  understand the scene and identify motion rule.
To achieve this goal, we design a text prompt about the question of scene type and real-world rule, and feed it into the VLM to obtain accurate scene understanding and motion analysis, according to  the object information obtained by object perception module. 

\noindent\textbf{Reasoning Object Relationships.}
In this stage, the VLM model utilizes the integrated information from the previous stage to infer the relationships between different objects according to the spatial arrangement, sizes, types. By analyzing the relationships, it aids to identify interaction possibility and interaction reasoning between objects. This reasoning provides essential support for the following interaction trajectory  reasoning. 

\noindent\textbf{Interaction Trajectories Reasoning.}
The model first identifies the types of interactions between objects, which can be categorized into three typical types: collisions and chain reactions, gravity and forces, and levers and mirrors. Then, based on the input motion trajectory of controlled object, the VLM predicts the trajectories of objects  affected by controlled object. 

\noindent\textbf{Iterative Reasoning and Ranking.}
This stage aims to refine the initial trajectory prediction of all objects. 
Following the initial trajectory prediction, the model engages in iterative reasoning to refine the results. By ranking the outputs of multiple iterations, the VLM identifies most plausible sequences of motion and collision responses, ensuring the consistency and accuracy of the reasoning process.

\noindent\textbf{Validation and Final Reasoning Outcome.}
Finally, we perform a forward and backward validation about predicted trajectories. The forward validation aims to verify whether the predicted trajectories adhere to the scene rules and context. The backward validation infers the input trajectories and object movements from the final generated trajectories through reverse engineering, then compare them with the actual inputs. Any discrepancies trigger re-iterations, leading to adjustments in the reasoning  until the most accurate and reliable final results are obtained.

\vspace{-8pt}
\section{The VOI Dataset}

To better evaluate the performance of trajectory-based video generation, we build a new video object interaction (VOI) dataset,  mostly collected from existing datasets such as LaSOT~\cite{fan2019lasot} and CADP~\cite{shah2018cadp}. Table~\ref{table:dataset} provides the details of our built VOI dataset, which contains 72 videos belonging to three different types.

\subsection{Data and Annotation}

The VOI dataset can be mainly divided into three subsets: \textit{collision and chain reaction}, \textit{gravity and forces}, and \textit{levers and mirrors}. \textit{(i) Collision and Chain Reaction.}
    This subset includes scenes involving collisions between objects and subsequent chain reactions, for instance, billiard balls colliding on a pool table. The subset can validate the ability of model to deal with complex interactions between objects.
\textit{(ii) Gravity and Force.}
    In this subset, the focus is on objects interacting under the influence of gravity and external forces. Examples include the free fall of basketballs, and objects controlled by human actions like throwing or kicking. The goal is to assess how well models understand the effects of gravitational forces and how external forces alter the motion of objects.
\textit{(iii) Levers and Mirrors.}
    This subset focuses on coupled systems, such as levers or mirror reflections. Examples include seesaws and objects that interact with mirrors (such as people checking their own reflections). It aims to assess how the model handles reflections, rotations, and complex motion patterns produced by lever systems or reflective surfaces.

\begin{table}[t]
\centering
\footnotesize
\setlength{\tabcolsep}{0.9mm}
\centering
\begin{tabular}{|cc|c|c|c|}
\hline
\multicolumn{2}{|c|}{Category} & Video & Anno Boxes & Anno Trajectories \\ \hline
\multicolumn{1}{|c|}{\multirow{3}{*}{\begin{tabular}[c]{@{}c@{}}Collision and \\ Chain Reaction\end{tabular}}}& Billiard & 16 & 2160 & 198 \\ \cline{2-5} 
\multicolumn{1}{|c|}{} & NewtonCradle & 7 & 300 & 79 \\ \cline{2-5} 
\multicolumn{1}{|c|}{} & Traffic & 10 & 180 & 90 \\ \hline
\multicolumn{1}{|c|}{\multirow{2}{*}{\begin{tabular}[c]{@{}c@{}}Gravity and \\ Force\end{tabular}}} & Basketball & 6 & 1080 & 34 \\ \cline{2-5} 
\multicolumn{1}{|c|}{} & FootBall & 7 & 960 & 76 \\ \hline
\multicolumn{1}{|c|}{\multirow{2}{*}{\begin{tabular}[c]{@{}c@{}}Levers and \\ Mirrors\end{tabular}}} & Seesaw & 15 & 840 & 145 \\ \cline{2-5} 
\multicolumn{1}{|c|}{} & Mirror & 11 & 1800 & 89 \\ \hline
\multicolumn{1}{|c|}{Total} & - & 72 & 7320 & 711 \\ \hline
\end{tabular}
\caption{An overview of our proposed VOI dataset. This dataset has 72 videos and contains three typical types of object interactions, including \textit{collision and chain reaction}, \textit{gravity and force}, and \textit{levers and mirrors}. We counted the number of videos, annotated boxes, and the objects trajectories.}
\label{table:dataset}
\vspace{-6pt}
\end{table}
\noindent\textbf{Annotations.}
For quantitative evaluation, we provide the bounding-box and trajectory annotations of objects. Specifically, for the videos having object bounding-box annotations (obtained from existing datasets), we directly use the original bounding-box annotations. For the videos without  annotations, we use the open-set detector, GroundingDINO \cite{liu2023grounding}, to identify objects  and obtain object bounding-boxes. Inspired by \cite{wu2024draganything}, we use the cotracker algorithm to track center points of objects to obtain accurate drag trajectories. Finally, we manually verify the annotated trajectories to ensure that annotations are accurate. As shown in Table~\ref{table:dataset}, there are 711 annotated trajectories among 72 videos. More details about VOI dataset are presented in suppl. material.

\subsection{Evaluation Metric}

Currently, performance evaluation in video generation primarily relies on  Fréchet Inception Distance (FID) ~\cite{Seitzer2020FID} and Fréchet Video Distance (FVD) ~\cite{unterthiner2018FVD}, where FID measures visual quality and FVD evaluates temporal consistency. However, these metrics do not assess the quality of object motion trajectories. To address this limitation, ObjMC~\cite{wang2024motionctrl} has been proposed to measure the similarity between the predicted and ground-truth trajectories of controlled objects, which does not consider other affected objects. To evaluate performance in more complex scenarios, we introduce MOC (\textbf{M}oving  \textbf{O}bject \textbf{C}onsistency), which measures similarity between predicted and ground-truth trajectories of all moving objects. The MOC is computed as 
\begin{equation}
MOC =\frac{1}{N} \sum_{n=1}^N \sqrt{\left(x_n^{\text {p}}-x_n^{\text {gt}}\right)^2+\left(y_n^{\text {p}}-y_n^{\text {gt}}\right)^2},
\end{equation}
where \( (x^{\text{p}}, y_t^{\text{p}}) \) and \( (x^{\text{gt}}, y_t^{\text{gt}}) \) are the predicted and real object positions. $N = \sum_{t=1}^{N_v} N_{t}^f*N_{t}^o$, where $N_v$ is the number of videos, $N_t^f$ is the number of frames at video $t$, and $N_t^o$ is the number of objects at video $t$.

\section{Experiments}

\begin{table}[t]
\centering
\footnotesize
\setlength{\tabcolsep}{4.5mm}
\begin{tabular}{l|ccc}
\hline
Methods   & FVD$\downarrow$ & FID$\downarrow$ & MOC$\downarrow$  \\ 
\hline
PhysGen~\cite{liu2024physgen} & 1795.87 & 129.30 & 118.36 \\
DragAnything~\cite{wu2024draganything} & 1295.94 & 129.07 & 71.50 \\
DragNUWA~\cite{yin2023dragnuwa}& 1034.89 & 102.80 & 72.09 \\
\rowcolor{myColor}
\textbf{C-Drag (Ours)}    & 771.83 & 95.87 & 46.47      \\ \hline
\end{tabular}
\caption{Comparison with existing works on the entire VOI dataset. Our C-Drag exhibits superior performance  in video quality (FVD), image quality (FID), and object motion consistency (MOC) with consistent gains compared to existing methods.}
\label{table:performance}
\end{table}

\begin{table}[]
\centering
\footnotesize
\setlength{\tabcolsep}{0.8mm}
\begin{tabular}{c|l|cccc}
\hline
 &  & Collision and & Gravity and & Levers and &  \\
\multirow{-2}{*}{Metric} & \multirow{-2}{*}{Method} & Chain Reaction & Force & Mirrors  \\ \hline
 & Physgen~\cite{liu2024physgen} & 1902.75 & 3039.89 & 2992.23  \\
 & DragAnything~\cite{wu2024draganything} & 1823.35 & 2152.93 & 1902.90  \\
 & DragNUWA~\cite{yin2023dragnuwa} & 1179.98 & 2318.95 & 1721.08  \\
\multirow{-4}{*}{FVD $\downarrow$} & \myCellColor \textbf{C-Drag (Ours)} & \myCellColor 1155.33 & \myCellColor 1218.15 & \myCellColor 1162.18  \\ \hline
 & Physgen~\cite{liu2024physgen} & 121.91 & 115.90 & 145.38  \\
 & DragAnything~\cite{wu2024draganything} & 114.38 & 139.36 & 142.56  \\
 & DragNUWA~\cite{yin2023dragnuwa} & 76.53 & 113.38 & 130.86  \\
\multirow{-4}{*}{FID $\downarrow$} & \myCellColor \textbf{C-Drag (Ours)} & \myCellColor 71.00 & \myCellColor 108.89 & \myCellColor 120.92  \\ \hline
 & Physgen~\cite{liu2024physgen} & 169.02 & 124.05 & 94.81  \\
 & DragAnything~\cite{wu2024draganything} & 94.95 & 101.96 & 45.39  \\
 & DragNUWA~\cite{yin2023dragnuwa} & 99.37 & 89.52 & 57.33  \\
\multirow{-4}{*}{MOC $\downarrow$} & \myCellColor \textbf{C-Drag (Ours)} & \myCellColor 73.52 & \myCellColor 44.10 & \myCellColor 22.64  \\ \hline
\end{tabular}
\caption{Comparison with existing works on VOI three subsets.  Our C-Drag outperforms existing methods on all subsets.}

\label{table:scenarios_performance}
\vspace{-6pt}
\end{table}

\subsection{Implementation Details and Metrics}

\noindent\textbf{Implementation Details.}
The experiments are performed on our VOI dataset.  We first employ our plug-and-play, training-free approach to generate object trajectories, and feed these generated trajectories to DragNUWA \cite{yin2023dragnuwa}. In the object perception module, we use segment anything model (SAM)~\cite{kirillov2023segment} as the class-agnostic segmentation method and GroundingDINO~\cite{liu2023grounding} as the open-set detection method. We directly adopt the pre-trained DragNUWA to generate videos without any re-training. We implement our proposed approach on a single NVIDIA RTX A6000 GPU.

\noindent \textbf{Evaluation Metrics.} We adopt the metrics of FID and FVD to  assess the quality of generated videos. Moreover, we introduce to use MOC to evaluate the motion consistency of all moving objects across frames. 

\begin{figure*}[ht]
  \centering
  \includegraphics[width=1\linewidth]{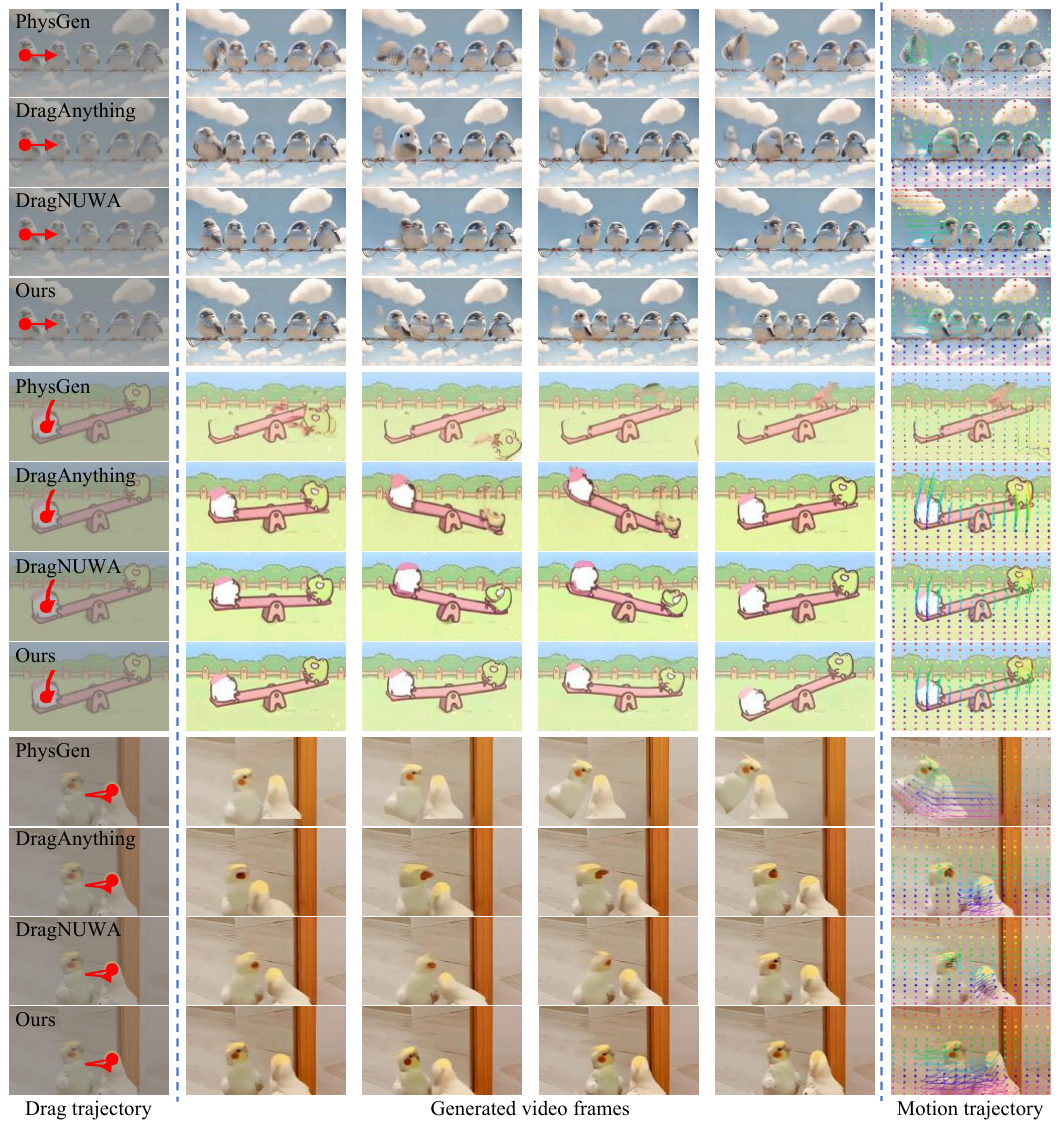}
    \caption{Qualitative comparison of our C-Drag with existing methods. PhysGen~\cite{liu2024physgen} (Rows 1, 5, 9) struggles with deformable objects and non-planar scenarios, requiring extensive manual parameter tuning, which leads to unrealistic movements in complex scenes. For example, in Row 5, the character falls since the seesaw boundary is not set, causing incorrect interactions. Similarly, both DragAnything~\cite{wu2024draganything} (Rows 2, 6, 10) and DragNUWA~\cite{yin2023dragnuwa} (Rows 3, 7, 11) have issues 
   when uncontrolled objects lose temporal consistency, such as merged birds in Rows 2 and 3, and severely deformed mirror-reflected objects in Rows 10 and 11. 
    In contrast, our C-Drag not only perceives and infers the movements of all objects but also maintains temporal consistency for all elements. Additional results are presented in suppl. material. }
  \label{fig:comparison2}
  \vspace{-6pt}
\end{figure*}

\subsection{Comparison With Existing Methods}

Table~\ref{table:performance} compares our proposed C-Drag with some existing methods on entire VOI dataset. We implement all these methods using their source codes, and report the results of FID, FVD, and MOC. Compared to existing methods, our proposed method has better performance. For instance, Physgen \cite{liu2024physgen} achieves the FVD score of 1795.87, DragNUWA \cite{yin2023dragnuwa} achieves the FVD score of 1034.89, while our method has the FVD score of 771.83. Therefore, our proposed method outperforms Physgen and DragNUWA by 1024.04
 and 263.06, demonstrating that our proposed method can generate the videos with higher quality. On MOC, DragAnything \cite{wu2024draganything} has the score of 71.50, DragNUWA has the score of 72.09, while our method has the score of 46.47. Namely, our proposed method outperforms DragAnything and DragNUWA by 25.03
 and 25.62, which demonstrates that our method can generate more accurate trajectories for all moving objects. 
  In addition, we present the results on three subsets in Table \ref{table:scenarios_performance}. Our proposed method outperforms existing methods on all three subsets. \\
\noindent\textbf{Qualitative Comparison.}
Figure~\ref{fig:comparison2} presents a qualitative comparison of our C-Drag with PhysGen, DragAnything, and  DragNUWA. The input image and drag trajectory are shown in first column. We select example frames from the generated videos, and present them from column 2 to column 5. We also show the motion trajectory obtained by CoTracker~\cite{karaev23cotracker} at last column. In Rows 1-4, our C-Drag accurately generates videos with the bird motion under crowded scene. In Rows 5-8, C-Drag predicts the seesaw motion which is controlled by gravity. In Rows 9-12, C-Drag generates the bird along with inside the mirror at the same time.

\begin{table}[t]
    \centering
    \footnotesize
    \setlength{\tabcolsep}{3.5mm}
    \begin{tabular}{cc|ccc}
    \hline
    OPM & CoT-Reasoning & FVD$\downarrow$ & FID$\downarrow$ & MOC$\downarrow$ \\ \hline
     &  & 1034.89 & 102.80& 72.09 \\
     & \checkmark  & 955.61 & 100.49 & 67.13 \\
    \rowcolor{myColor} 
    \checkmark  & \checkmark  & 771.83 & 95.87 & 46.47 \\ \hline
    \end{tabular}
    \caption{Impact of the two modules in our C-Drag. OPM represents object perception module, and CoT-Reasoning represents CoT-based motion reasoning module. The best results are obtained when integrating both modules into the baseline.}
    \label{table:ablationmodule}
    \vspace{-6pt}
\end{table}

\begin{table}[t]
    \centering
    \footnotesize
    \setlength{\tabcolsep}{1.4mm}
    \begin{tabular}{ccc|ccc}
    \hline
     VLM & OS Detection & CA Segmentation & FVD$\downarrow$ & FID$\downarrow$ & MOC$\downarrow$ \\ \hline
       &  &  & 1034.89 & 102.80& 72.09 \\
     \checkmark  &  &  & 921.05 & 100.13 & 63.84 \\
     \checkmark  & \checkmark  &  & 782.25 & 96.74 & 49.53 \\
    \rowcolor{myColor}
     \checkmark  & \checkmark  & \checkmark  & 771.83 & 95.87 & 46.47 \\ \hline
    \end{tabular}
    \caption{Effect of different components (VLM, open-set detection, class-agnostic segmentation) in object perception module.}
    \label{table:sub-OPM}
    \vspace{-12pt}
\end{table}

\subsection{Ablation Study}

Here, we report our extensive ablations and report the results  on entire VOI dataset.

\noindent\textbf{Effect of Different Modules.}  Table~\ref{table:ablationmodule} shows the results of integrating the object perception module (OPM) and CoT-based motion reasoning module into the baseline. We note that the baseline is DragNUWA \cite{yin2023dragnuwa}. When only integrating CoT-based reasoning into the baseline, it has 79.28 improvement on FVD, 2.31 improvement on FID, and 4.96 improvement on MOC. When further adding OPM, it has 183.78 improvement on FVD, 4.62 improvement on FID, and 20.66 improvement on MOC.

\begin{table}[t]
    \centering
    \footnotesize
    \setlength{\tabcolsep}{2.7mm}
    \begin{tabular}{ccccc|ccc}
    \hline
    S1 & S2 & S3 & S4 & S5 & FVD$\downarrow$ & FID$\downarrow$ & MOC$\downarrow$ \\ \hline
      &  &  &  & \multicolumn{1}{c|}{} & 1034.89 & 102.80& 72.09 \\
    \checkmark  &  &  &  & \multicolumn{1}{c|}{} & 996.55 & 101.61 & 70.04 \\
    \checkmark  & \checkmark  &  &  & \multicolumn{1}{c|}{} & 983.42 & 99.14 & 67.02 \\
    \checkmark  & \checkmark  & \checkmark  &  & \multicolumn{1}{c|}{} & 897.11 & 98.25 & 62.35 \\
    \checkmark  & \checkmark  & \checkmark  & \checkmark  & \multicolumn{1}{c|}{} & 789.13 & 96.21 & 53.86 \\
    \rowcolor{myColor} 
    \checkmark  & \checkmark  & \checkmark  & \checkmark  & \multicolumn{1}{c|}{\checkmark} & 771.83 & 95.87 & 46.47 \\ \hline
    \end{tabular}
    \caption{Effect of different stages in CoT-based motion reasoning module. S1: Scene and Object Understanding; S2: Reasoning Object Relationships; S3: Interaction Trajectories Reasoning; S4: Iterative Reasoning and Ranking; S5: Validation and Final Reasoning Outcome.}
    \label{table:diff-step}
    \vspace{-6pt}
\end{table}

\noindent\textbf{On Different Components of Object Perception.}
Our object perception module employs a VLM, open-set detector and a class-agnostic segmentation component to extract objects from input image. 
Table~\ref{table:sub-OPM} presents impact of these components. When only using VLM, we achieve FVD of 921.05. The results improve by introducing open-set detection and a further improvement is obtained by integrating class-agnostic segmentation with FVD of 771.8.

\noindent\textbf{On Different stages of CoT-based Motion Reasoning.} 
Our CoT-based motion reasoning contains five stage reasoning. Table~\ref{table:diff-step} presents the impact of these stages. It can be observed that these stages gradually improve the results. The best results are obtained when using all five stages. 

\section{Conclusion}

We propose C-Drag, a training-free motion controller, that strives for improved video generation capabilities in multi-object interaction scenarios. Our C-Drag contains an object perception module and a CoT-based motion reasoning module. The object perception module aims to extract all objects within image, whereas the CoT-based motion reasoning module aims to predict the motion trajectories of all objects. To evaluate the generation performance on multi-object iteration scenes, we build a new dataset, VOI, and annotate the object ground-truth trajectories. Experiments on VOI dataset demonstrates that C-Drag generates videos with high motion consistency, compared to existing methods. While achieving promising performance, we observe C-Drag to occasionally struggle in case of complex trajectories involving multiple direction changes. A future direction is to explore integrating temporal modeling capabilities to capture temporal dynamics of complex trajectories.

{
    \small
    \bibliographystyle{ieeenat_fullname}
    \bibliography{main}
}


\end{document}